\documentclass[11pt, letterpaper]{article}
\usepackage[utf8]{inputenc}
\usepackage[margin=1.5cm]{geometry}
\usepackage{titlesec}
\usepackage{tabu}
\usepackage{enumitem}
\usepackage{amssymb}
\usepackage{xcolor}
\newlist{selectlist}{itemize}{2}
\setlist[selectlist]{label=$\square$,leftmargin=*,noitemsep,topsep=0pt}
\usepackage{comment}
\usepackage{amsmath,graphicx}
\usepackage{amsfonts}
\usepackage{lmodern}
\usepackage{etoolbox}
\patchcmd{\thebibliography}{\section*{\refname}}{}{}{}
\usepackage{algorithm}
\usepackage[noend]{algpseudocode}
\usepackage{hyperref}
\hypersetup{
    colorlinks=true,
    linkcolor=blue,
    filecolor=magenta,      
    urlcolor=blue,
}
 
\titleformat{\section}[block]{\hspace{1em}\bfseries}{\thesection.}{0.5em}{} 
\titleformat{\subsection}[block]{\hspace{1em}}{\thesubsection}{0.5em}{}

\begin{document}

\noindent
\textbf{\textit{CP-AGCN: Pytorch-based Attention Informed Graph Convolutional Network for Identifying Infants at Risk of Cerebral Palsy}}
\vskip0.5cm
\noindent
\textbf{\textit{Haozheng Zhang (Durham University, haozheng.zhang@durham.ac.uk), \\
Edmond S. L. Ho (University of Glasgow, Shu-Lim.Ho@glasgow.ac.uk), \\
Hubert P. H. Shum (Durham University, hubert.shum@durham.ac.uk, corresponding author)}}\\

\noindent
\textbf{Abstract}\\
\textit{
Early prediction is clinically considered one of the essential parts of cerebral palsy (CP) treatment. We propose to implement a low-cost and interpretable classification system for supporting CP prediction based on General Movement Assessment (GMA). We design a Pytorch-based attention-informed graph convolutional network to early identify infants at risk of CP from skeletal data extracted from RGB videos. We also design a frequency-binning module for learning the CP movements in the frequency domain while filtering noise. Our system only requires consumer-grade RGB videos for training to support interactive-time CP prediction by providing an interpretable CP classification result. 
}
\vskip0.5cm

\noindent
\textbf{Keywords}\\
\textit{Disease prediction; Graph convolutional network; Classification; Human motion analysis}
\vskip0.5cm
\newpage
\noindent
\textbf{Code metadata}\\

\noindent
\begin{tabular}{|l|p{6.5cm}|p{9.5cm}|}
\hline
\textbf{Nr.} & \textbf{Code metadata description} & \textbf{Please fill in this column} \\
\hline
C1 & Current code version & v1 \\
\hline
C2 & Permanent link to code/repository used for this code version & 
\href{https://github.com/zhz95/CP-AGCN}{https://github.com/zhz95/CP-AGCN} \\
\hline
C3  & Permanent link to Reproducible Capsule & \href{https://codeocean.com/capsule/6073072/tree/v1}{https://codeocean.com/capsule/6073072/tree/v1} \\
\hline
C4 & Legal Code License   & MIT License \\
\hline
C5 & Code versioning system used & git \\
\hline
C6 & Software code languages, tools, and services used & Python \\
\hline
C7 & Compilation requirements, operating environments \& dependencies & Python 3.7, PyTorth 1.8.10, OpenPose 1.7.0 \\
\hline
C8 & If available Link to developer documentation/manual & \href{https://codeocean.com/capsule/6073072/tree/v1}{https://codeocean.com/capsule/6073072/tree/v1} \\
\hline
C9 & Support email for questions & haozheng.zhang@durham.ac.uk \\
\hline
\end{tabular}\\

\vskip0.5cm
\noindent
\section{Introduction}
Due to the abnormal development or damage to parts of the brain, cerebral palsy (CP) appears in the early stages of the patient's childhood and permanently affects the patient's quality of life. There are about $2-3$ CP patients in 1000 children in the UK~\cite{Carter2019}, which is similar to other developed countries. Although CP cannot be completely cured at present, early prediction of CP and intervention are considered as a paramount part of the treatment.

Current clinical early prediction of CP is investigated by General Movement Assessment (GMA)~\cite{Einspieler2005}. GMA can be done in person by GM assessors to assess an infant, or it can be done via watching an RGB video that has recorded the general movements of the infant. However, the GMA training is time- and resource-consuming, making it challenging to cope with the high demand for CP prediction. To tackle this problem, we propose automating this process by analyzing the general movements of infants from RGB videos. This allows the early prediction to cover even the lower-risk population. Motivated by the encouraging results reported in recent research based on skeletal data \cite{McCay2019,Sakkos2021,McCay2019a,Nguyen2021,McCay:BHI2021,Zhu:BHI2021,McCay2022,PCPP}, the 2D joint locations of the infant are extracted from RGB videos as the input of the system for CP prediction. The computational intelligence of our system is implemented with a graph convolution network, a kind of deep artificial neural network that models relational data very well, making it suitable for skeleton data. 
Our graph convolution operation encodes the input of human joint movement as frequency features in the graph structure. In addition, we employed the attention mechanism to enhance the learning performance with better interpretability. As a result, the implemented system can model the graph structure of the human pose features and amplify the influence of the most important parts of the data that are considered by the neural network.

We implement our system in PyTorch due to the availability of compatible open-source resources and its good coverage of the required deep learning functionalities. Firstly, we adapt OpenPose~\cite{cao2019openpose}, a pose estimation software, to extract the human skeletal pose features from only the consumer-level RGB videos. Then, we implement the Fast Fourier Transform (FFT) algorithm with a tailor-made binning operation to analyse the human joint movement frequencies from the pose features. With the frequency operations, we can reduce the noise in the data and filter the high-frequency movement features that are less relevant for the CP classification~\cite{Rahmati2016,wode, McCay2022}. We implement our CP classification GCN network based on the backbone (ST-GCN) framework from Shi \textit{et al.}~\cite{yan2018}, because it shows promising classification performance and is widely identified as the benchmark in human pose-based studies. We validate our system in two datasets, showcasing that our system achieves state-of-the-art performance. The theoretical background of the system has been published in~\cite{wode}. In this paper, we explain the software implementation details and discuss the impact of our software.

\section{System description}
\begin{figure}
\includegraphics[width=\textwidth]{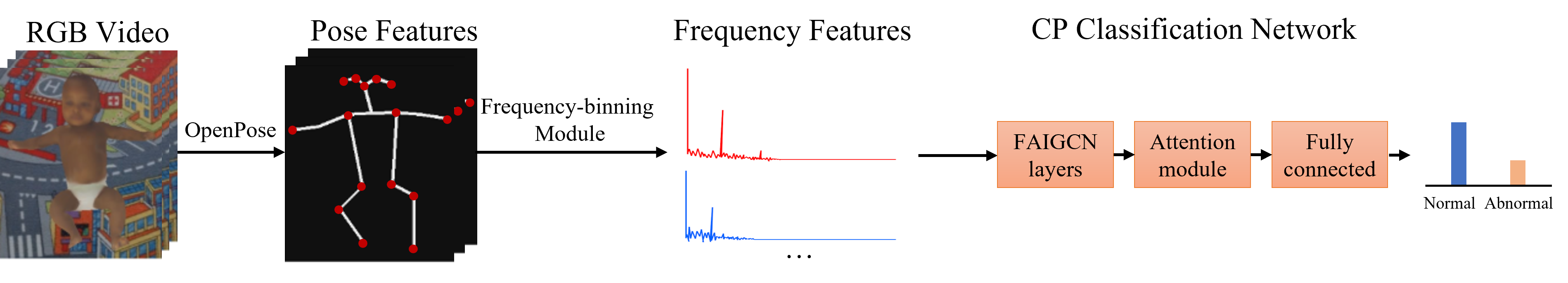}
\caption{The architectures of proposed CP classification system.}
\label{fig1}
\end{figure}

\begin{algorithm}
\caption{Attention GCN for CP classification}
\textbf{Inputs}: $\mathbf{H}$: the set of joint features (dim: $B\times N$); $A$: the adjacency matrix of the graph distance between joints (dim: $N\times N$); An one-hot classification label embedding.\\
\textbf{Output:} A $0/1$ classification result. 
\begin{algorithmic}[1]
\State \begin{math} \tilde{\mathbf{A}} = \mathbf{A}+\mathbf{I}_L \end{math}, \begin{math} \tilde{\mathbf{D}}_{ii} = \sum_j \tilde{\mathbf{A}}_{ij}  \end{math}.
\If{s $<$ S}:
\State Take the graph convolution operation by Eq. \ref{GCN}, where $\mathbf{W}$ is updated by Eq. \ref{agg} and \ref{h}.
\State s $\leftarrow$ s $+1$.
\EndIf
\State Encode the output to a 0/1 classification result by concatenation.
\end{algorithmic}
\label{algorithm1}
\end{algorithm}

As shown in Figure~\ref{fig1}, our proposed CP classification system first extracts infants' pose features (i.e., joint positions) by the pose estimation algorithm OpenPose~\cite{cao2019openpose}, due to its robust performance. The latest version of OpenPose (1.7.0) comes with the pre-trained model, such that one can use the network effectively without training. Then, we implement a frequency-binning module to convert the pose features into the Fast Fourier Transform (FFT) coefficients for learning CP in the frequency domain. Both pose features or frequency features are fed into our designed PyTorch-based attention-informed graph convolutional network (CP-AGCN) for classifying normal and abnormal CP labels. 

In particular, the frequency-binning module adapts Bluestein’s FFT algorithm~\cite{Bluestein1970} for feature conversion. We tailor-made a binning method to filter high-frequency components, which mainly consist of data noise and high-frequency movement features:
\begin{equation}
f_{n} = \begin{cases}
         $Round$( f_{0} \cdot c^n ), &\text{if $f_{n} \cdot c^n< 3$,}\\
         $Ceiling$( f_{0} \cdot c^n ), &\text{if $f_{n} \cdot c^n \geq 3$,}
        \end{cases}
        \label{eqbin}
\end{equation}
where $f_n$ ($f_0=1$) is the width of the $n^{th}$ bin, and $c$ is a parameter that control the bin width. 

The design in Eq.~\ref{eqbin} enables the classification network to focus on learning the low-to-mid band movement frequency features. To retain the low-to-mid frequency human movement information, this function takes the round for widths less than three units. This function can distinguish the middle frequency and high-frequency bands based on the exponential growth of $c$ and the ceiling function. Since we have not included any learnable parameter in this module, such a low computational cost module is adaptable for both machine learning-based or deep neural network (DNN) based classification systems with 24-60 FPS input videos.

For CP classification, the algorithm flow is shown in Algorithm \ref{algorithm1}. We employ the GCN~\cite{Kipf2017} as the benchmark model to learn the joint dependencies from the skeleton graph. The graph convolutional propagation is demonstrated by:
\begin{equation}
    \mathbf{H}^{(l+1)} = \sigma \left(\tilde{\mathbf{D}}^{-\frac{1}{2}}\tilde{\mathbf{A}}\tilde{\mathbf{D}}^{-\frac{1}{2}}\mathbf{H}^{(l)} \mathbf{W}^{(l)} \right)
    \label{GCN}
\end{equation}    
where \begin{math} \tilde{\mathbf{A}} = \mathbf{A}+\mathbf{I}_L \end{math} is the adjacency matrix. \begin{math} \tilde{\mathbf{D}}_{ii} = \sum_j \tilde{\mathbf{A}}_{ij}  \end{math}. \begin{math} \mathbf{I}_L  \end{math} is an identity matrix with \textit{L} dimensions. \begin{math} \mathbf{W}^{(l)}\end{math} is the learnable weight matrix of layer $l$. \begin{math} \sigma(\cdot) \end{math} is the \textit{ReLU} nonlinear activation function.

By using the architecture backbone from~\cite{yan2018}, we apply the human skeleton graph \begin{math} G = (V,E)\end{math} for interpreting the importance ranking of joint's features CP classification. In this graph, \begin{math} \{V = {v_{bi}|b=1,...,B; i=1,...,N}\}\end{math} is the set of joint features, where \begin{math} v_{bi} \end{math} is the \textit{b}-th bin of \begin{math} i\end{math}-th joint. The edge set \textit{E} consists of: (1) the intra-skeleton connection, \begin{math} \{v_{bi}v_{bj}|(i,j)\in K \} \end{math}, where \begin{math} K \end{math} is the amount of the natural connections of skeleton joints. (2) the inter-feature edges which connect the bins of a joint from low to high, \begin{math} \{v_{bi}v_{(b+1)i}\} \end{math}. 

We propose to use attention-informed mechanism to learn the weight of features. We aggregate the network input features \begin{math}\{ \mathbf{h}_{1,i},\mathbf{h}_{2,i},\dots,\mathbf{h}_{B,i} \}\end{math} with attentions \begin{math}\alpha _{b,i}\end{math} by:
\begin{equation}
    \mathbf{v}_k = \sum_{b=1}^{B} \alpha_{b,i}\mathbf{h}_{b,i}
    \label{agg}
\end{equation}
and the attention weight \begin{math}\alpha _{b,i}\end{math} is calculated by:
\begin{equation}
    \alpha_{b,i} = \frac{\text{exp} \left(\sigma^{'}_n\left(\mathbf{w}_{\alpha}^\mathsf{T},  \mathbf{z}_{b,i}\right)\right)}{\sum_{b}\text{exp}\left(\sigma^{'}\left( \mathbf{w}_{\alpha}^\mathsf{T},\mathbf{z}_{b,i}\right)\right)},
    \quad \quad
    \mathbf{z}_{b,i} = tanh\left( \mathbf{W}_{z}\mathbf{h}_{b,i}\right)
    \label{h}
\end{equation}
where $\sigma^{'}_n$ is an designable activation function
$ \sigma^{'}_n =\mathbf{w}_{\alpha}^\mathsf{T}\mathbf{z}_{\alpha} 
$, and $\mathbf{w}_{\alpha}$ and $\mathbf{W}_{z}$ are learnable parameters.

The introduced system above is programmed under the PyTorch framework with packages including numpy 1.20.3, torch 1.8.1 and torchvision 0.3.0. We conduct comprehensive experiments on a synthetic dataset (MINI-RGBD)~\cite{hesse2018b} with annotation provided by~\cite{McCay2019a} and RVI-38 dataset~\cite{McCay2022}, the result shows our system achieves state-of-the-art performance on both datasets~\cite{wode}.

Regarding computational costs, although we have applied the attention mechanism, our low-cost system can achieve a training speed of around 4 frames/second with an NVIDIA GeForce RTX 3080, with only 0.3 frames/second drop compared with our variant without the attention module. It means the total system training time on a 12 video sequences (1000 frames each) dataset is only about 50 minutes, including OpenPose pose estimation. This is considered to be very efficient in deep learning algorithms. During run-time, it only needs about 45s for the CP classification of a 33.3s 30FPS video with an Intel Core i7 CPU (i.e., no GPU needed). It shows that our system is employable for interactive-time prediction in a daily hospital environment with normal computer equipment setting, and is a feasible solution to support CP early prediction. 

\section{Impact overview}
Our proposed system is expected to provide lower-risk patients with low-cost, non-intrusive CP abnormal movement classification results as a warning sign, with its accuracy supported by two datasets \cite{wode}. This provides a way for supporting the early prediction of CP in the clinical resource-limited areas, and relieving the labour stress of expert-based GMA \cite{Einspieler2005}. In addition, our system can provide clinicians with information about the importance-ranking of joints' movement or frequency features in CP classification by the attention weights. 

Apart from the early prediction of CP, we expect this system can be extended for a number of human motor disorders. For example, Parkinson's disease patients usually have motor dysfunction symptoms like tremors, unpaired gait patterns and limb stiffness \cite{poewe17parkinson}. The movement frequency of patients is highly likely to be different from that of healthy people, particularly due to tremors, and the use of frequency features may inform deep learning based Parkinson's disease diagnostic systems \cite{zhang22posebased}. 

This easy-to-use system also benefits researchers in different fields like body motion and hand gesture analysis since it provides an alternative way for learning these features in the frequency domain. Recent work has shown that hand tremor can be modelled and reduced by deep learning as a denoising process \cite{leng21image}, but the use of frequency is still yet to be considered. Similarly, frequency information is likely to be useful for sports motion analysis and visualisation \cite{shen17posture}. In addition, the users can extend our system for multi-class classification tasks by minor modifications on the final fully connected layer. Existing work already shows that frequency information is useful for action recognition \cite{8897586} and gesture recognition \cite{670984}. Our frequency binning idea and its uses on a graph neural network may inform such applications.

\section{Conclusion and future improvements}
In this work, we implement a Pytorch-based attention-informed graph convolutional network to classify cerebral palsy infants. We propose a frequency binning module for CP joint position features to increase to improve the classification performance, which is adaptable for other deep learning or traditional machine learning based classification models. Besides, we design an attention module to both improve the classification performance and interpret the joint features' importance ranking. Our system is validated on two datasets in~\cite{wode} and achieves interactive speed on a consumer-grade computer. 

In future versions, we consider integrating our system into an embedded system, such as the pixel processor array that integrates convolutional neural network operations \cite{9010705}, for autonomous disease monitoring and real-time decision-making. In addition, the performance of our system relies on the accuracy of the pose estimation process. Future versions will focus on improving the pose estimation part by adapting more advanced systems such as \cite{fang2017rmpe}. Finally, we wish to improve hardware compatibility by generalising the system with different cameras (e.g., unfocused camera, blurry camera). This can be implemented with a hardware abstraction layer that deals with device-specific procedures~\cite{abdelkader2017}, which connects to a unified prediction algorithm.

\hspace*{\fill} 
\\
\noindent 
\textbf{Declaration of competing interest}\\
The authors declare that they have no known competing financial interests or personal relationships that could have appeared to influence the work reported in this paper.


\hspace*{\fill} 
\\

\end{document}